\documentclass[manuscript]{acmart}

\usepackage{cite}
\usepackage{amsmath,amssymb,amsfonts}
\usepackage{graphicx}
\usepackage{textcomp}
\usepackage{xcolor}
\usepackage{algpseudocode}
\usepackage{algorithm}
\usepackage[T1]{fontenc}
\usepackage{amsmath}
\usepackage{esvect}
\usepackage{caption}

\copyrightyear{2025}
\acmYear{2025}
\setcopyright{rightsretained}
\acmConference[CSAIDE 2025]{2025 4th International Conference on Cyber Security, Artificial Intelligence and the Digital Economy}{March 07--09, 2025}{Kuala Lumpur, Malaysia}
\acmBooktitle{2025 4th International Conference on Cyber Security, Artificial Intelligence and the Digital Economy (CSAIDE 2025), March 07--09, 2025, Kuala Lumpur, Malaysia}
\acmPrice{}
\acmDOI{10.1145/3729706.3729819}
\acmISBN{979-8-4007-1271-5/25/03}
\begin{document}

\title{Context-Based Fake News Detection using Graph Based Approach: A COVID-19 Use-case}

\author{Chandrashekar Muniyappa}
\authornotemark[1]
\email{cnachiketa07@gmail.com}
\affiliation{%
  \institution{Independent Researcher}
  \city{Dublin}
  \country{USA}}

\author{Dr. Sirisha Velampalli}
\email{sirisha.velampalli@gmail.com}
\affiliation{%
  \institution{UCEK, JNTU Kakinada, LTIMindTree}
  \city{Hyderabad}
  \country{India}}
\renewcommand{\shortauthors}{Muniyappa et al.}

\begin{abstract}
In today's digital world, fake news is spreading with immense speed. Its a significant concern to address. In this work, we addressed that challenge using novel graph based approach. We took dataset from Kaggle that contains real and fake news articles. To test our approach we incorporated recent covid-19 related news articles that contains both genuine and fake news that are relevant to this problem. This further enhances the dataset as well instead of relying completely on the original dataset. We propose a contextual graph-based approach to detect fake news articles.  We need to convert news articles into appropriate schema, so we leverage Natural Language Processing (NLP) techniques to transform news articles into contextual graph structures. We then apply the \textit{Minimum Description Length (MDL)}-based \textit{Graph-Based Anomaly Detection (GBAD)} algorithm for graph mining. Graph-based methods are particularly effective for handling rich contextual data, as they enable the discovery of complex patterns that traditional query-based or statistical techniques might overlook. Our proposed approach identifies normative patterns within the dataset and subsequently uncovers anomalous patterns that deviate from these established norms.
\end{abstract}


\maketitle

\section{Introduction}
Graph-based representations have broad applications across various domains, including social networks, chemical compound analysis, large-scale network systems, semantic search, knowledge discovery, natural language processing, and cybersecurity. Traditional data mining techniques — such as classification, clustering, association analysis, and outlier detection — have been successfully adapted to graph mining, enabling the extraction of complex patterns and insights.

Conceptual Graphs are widely employed in domains like Natural Language Processing, Expert Systems, Database Design, and Information Retrieval Jesus A. Gonzalez et al.[1] These graphs act as an intermediate language that translates natural language data into structured, computer-oriented representations. Unlike conventional graph structures, conceptual graphs enable the modeling of both relationships and logical connections. In our study, conceptual graphs are utilized to model news articles. Once these graphs are constructed, we apply the Graph-Based Anomaly Detection (GBAD) algorithm to identify both normative (expected) and anomalous (potentially fake) patterns.

Fake news refers to verifiably false information disseminated through media channels, including fabricated articles, misinformation, and satirical content Xinyi Zhou et al. [3]. The widespread distribution of fake news poses a serious threat to democracy, journalism, and public trust. Consequently, effective detection mechanisms are crucial to mitigating its impact.

Existing fake news detection approaches leverage methods such as knowledge-based techniques, style analysis, propagation tracking, and source evaluation. Additionally, machine learning models, language-based techniques, topic-agnostic frameworks, and hybrid approaches have also been explored De Beer et al.[2]. In our work, we employ a novel graph-based approach that structures news articles into conceptual graphs. This method provides a meaningful structural representation of data, enhancing the discovery of complex patterns and insights that traditional techniques may overlook.

\textbf{Key Contributions:}
\begin{itemize}
    \item Modeling news articles as conceptual graphs for enhanced structural representation.
    \item To the best of our knowledge, this is the first attempt to apply the GBAD algorithm in the news domain for fake news detection.
    \end{itemize}
\section{Related Work}

Graph-based analytical approaches have been widely explored for anomaly detection in various domains. Velampalli et al., applied GBAD in the VAST Challenge 2017 to investigate the declining population of Rose-Crested Blue Pipit nesting birds. By identifying normative patterns in the park's ecosystem, the authors successfully reported anomalous behavioral patterns contributing to this decline Velampalli. S et al. [4]. Ramesh et al. studied various forms of anomalies — including temporal, spatial, and behavioral irregularities — using graph mining techniques. Their research utilized sensor data from smart homes to track residents' daily activities, enhancing the safety of elderly and cognitively impaired participants [5]. Chandra et al. proposed SAFER (Socially Aware Fake nEws detection fRamework), a graph-based model designed to detect fake news. The framework employed heterogeneous graph structures combined with a Graph Neural Network (GNN) model, outperforming traditional text-based methods [6]. Velampalli et al. leveraged NMap (Network Mapper) to collect system-level data from networked computers. They converted this data into conceptual graphs to identify potential vulnerabilities, outdated services, and unauthorized entities Velampalli. S et al.[9]. Paudel et al. applied GBAD with graph beam and MDL heuristics on the Kyoto dataset (CASAS program). Using sensor logs from 400 participants, they constructed activity graphs to uncover various anomalies [05]. Dong Zhou et al.[13] proposed Graph structure and guided semantic fusion technique and built a graph model to derive features across images and texts to identify fake news based on given context. Hiremath et al. [14] used Graph Neural Networks (GNN) along with Deep Neural Networks (DNN) to study scoial network data and identify fake news, by combining GNN with DNN they were able to achieve higher accuracy by reducing the false positive rate as graph embeddings can capture the relationship between entities better. Bingbing et al.[15] designed a novel heterogeneous graph with different types of nodes representing news, topics, and entities to capture the relationships between them, they used a Knowledge Graph (KG) to capture the factual information and applied attention mechanism to aggregate the information between them and achieved good results in identifying fake news. Benamira et al. [16], applied a graph-based semi-supervised learning method that achieved superior performance in identifying misinformation in networks. Teddy Lazebnik et al. [17], utilized temporal graph anomaly detection techniques to identify fake news by modeling time-evolving contextual relationships. Some of the recent studies have also explored novel graph approaches to identify fake news such as: AnomalyGFM: Graph Foundation Model for Zero/Few-shot Anomaly Detection [19].
Bots Shield Fake News: Adversarial Attack on User Engagement based Fake News Detection [20].
UniGAD: Unifying Multi-level Graph Anomaly Detection [21].
Context Correlation Discrepancy Analysis for Graph Anomaly Detection [22].
A Structural Information Guided Hierarchical Reconstruction for Graph Anomaly Detection [23]. Although conceptual graphs have seen extensive use across domains like security, finance, and healthcare, our work is the first to apply GBAD for detecting fake news in the news media domain. This novel application bridges the gap in utilizing graph-based techniques for uncovering misinformation patterns in news articles.

\section{Graph-Based Anomaly Detection (GBAD)}

\begin{figure}
\centering
\includegraphics[width=\columnwidth]{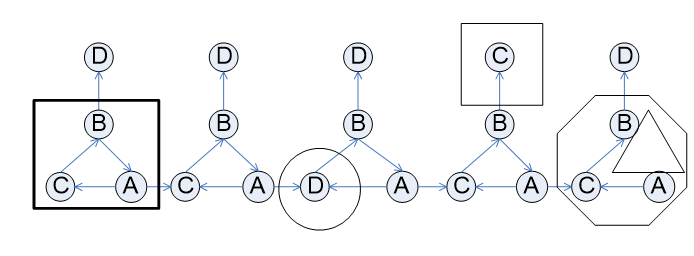}
\caption{Illustration of Different Types of Structural Anomalies Detected by GBAD}
\label{fig:gbad}
\end{figure}

In this study, we employ the Graph-Based Anomaly Detection (GBAD) system to identify both normative and anomalous patterns within graph-structured data. GBAD efficiently detects three types of structural anomalies: modifications, insertions, and deletions. Figure \ref{fig:gbad} illustrates these structural changes.

The GBAD system leverages three primary algorithms — GBAD-MDL, GBAD-P, and GBAD-MPS — to identify these anomalies. Each algorithm follows a distinct strategy for detecting unusual graph patterns. These algorithms are built upon the SUBDUE graph-based knowledge discovery system Nikhil S. Ketkar et al. [7] and employ the Minimum Description Length (MDL) principle [8] for pattern detection.

The MDL principle is a formalization of Occam's Razor, which states that the simplest explanation is usually the best one. In the context of GBAD, the MDL principle minimizes the combined complexity of the graph representation and its substructures using the following objective function:

\begin{equation}
M(S,G) = DL(G|S) + DL(S)
\end{equation}

Where:
\begin{itemize}
\item $G$ = The complete graph.
\item $S$ = The identified substructure.
\item $DL(G|S)$ = The description length of $G$ after compression using $S$.
\item $DL(S)$ = The description length of the substructure itself.
\end{itemize}

The GBAD system employs three distinct algorithms to detect anomalies:

\subsection{GBAD-MDL}
The GBAD-MDL algorithm identifies the best substructure in the graph using the MDL principle. Once the optimal pattern is identified, the algorithm searches for substructures that deviate slightly from this normative pattern, signaling potential anomalies. This approach is particularly effective in detecting modifications within the graph structure.

\subsection{GBAD-P}
The GBAD-P algorithm extends the MDL approach by examining all possible extensions to the normative substructure. It identifies graph patterns with extensions that have a lower probability of occurrence, thus highlighting unexpected structures. This algorithm is adept at detecting insertions within the graph.

\subsection{GBAD-MPS}
The GBAD-MPS algorithm identifies anomalies by detecting missing elements such as edges or vertices. The algorithm identifies these incomplete patterns by comparing them with the discovered normative structure. This method is particularly useful for detecting deletions within the graph.

Each of these algorithms provides a unique perspective on anomaly detection, allowing for a comprehensive analysis of graph-structured data. By leveraging these algorithms, GBAD can effectively identify a wide range of structural anomalies, making it a powerful tool for various applications.

\section{Graph Topology}

\begin{figure}[tb]
 \centering 
  \framebox{\includegraphics[width=9.0cm,height=5.0cm]{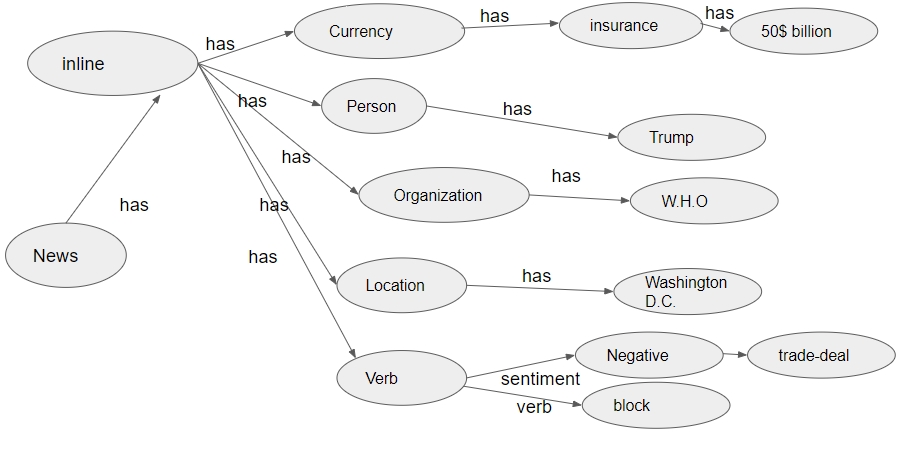}}
 \caption{Proposed Graph Topology}
 \label{fig:graph topology}
\end{figure}

We experimented with several candidate graph topologies, but found that the one shown in Figure \ref{fig:graph topology} provides the best results. We structure each news articles inline with Person, Organisation, Location etc., and the verb that contains positive and negative sentiment. We then used GBAD to discover normal as well as fake news articles. One instance of the input file obtained after pre-processing the data is shown in Table 1.

\begin{table}[]
\centering
\begin{tabular}{|l|}
\hline
\begin{tabular}[c]{@{}l@{}}XP \# 1\\ v 1 "News"\\ v 2 "in-line"\\ v 3 "Person"\\ v 4 "Organization"\\ v 5 "Location"\\ v 6 "Verb"\\ v 7 "Noun"\\ v 8 "Andres"\\ v 9 "congress"\\ v 10 "Mexico"\\ v 11 "infected"\\ v 12 "corona"\\ v 13 "president"\\ e 1 2 "has"\\ e 2 3 "has"\\ e 2 4 "has"\\ e 2 5 "has"\\ e 2 6 "has"\\ e 2 7 "has"\\ e 3 8 "has"\\ e 4 9 "has"\\ e 5 10 "has"\\ e 6 11 "has"\\ e 7 12 "has"\\ e 7 13 "has"\end{tabular} \\ \hline
\end{tabular}
\caption{Sample Instance}
\end{table}
\begin{figure}
\centering 
\label{fig:sample}
\end{figure}
We collected the real-news news articles from the "The New York Times" news paper and injected them into the real news kaggle dataset [18]. 
\begin{figure}[tb]
 \centering 
  \framebox{\includegraphics[width=9.4cm,height=6.8cm]{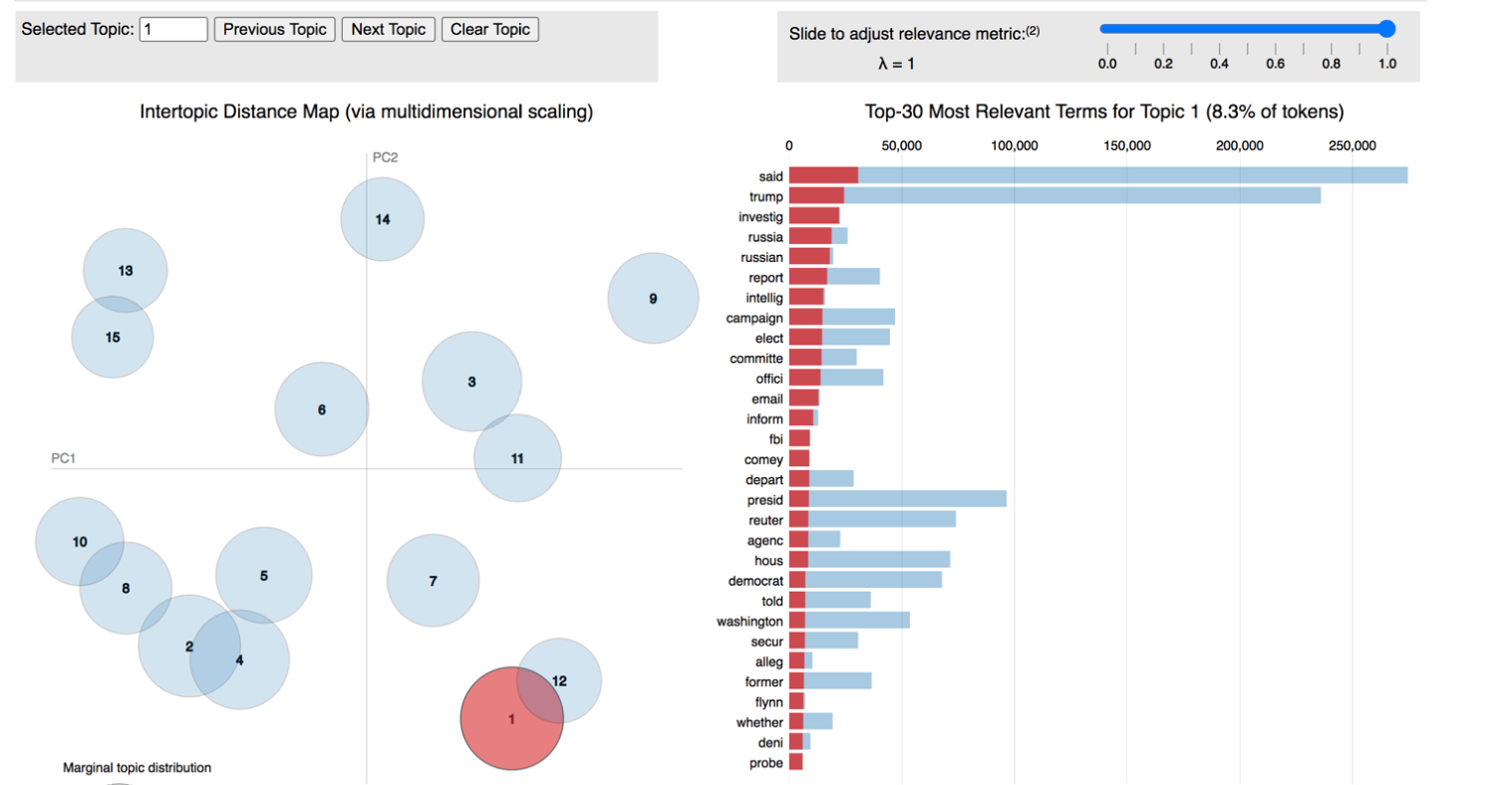}}
 \caption{Topics of Interest: Topic Modeling}
 \label{fig:topics1}
\end{figure}

Firstly, the news articles were cleaned using the NLP libraries and the LDA algorithm was applied for topic modeling, Blei et al. David M. Blei et al. [11] proposed “Latent Dirichlet Allocation” (LDA) algorithm that is based on Bayesian probabilities where topics will be extracted from the unstructured data by assigning probabilities to topics. This algorithm is widely used in unstructured data analysis and topic modelling. As part of this process totally 15 topics were identified as shown in Figure \ref{fig:topics1} out of these 15 topics, we picked only one topic "Politics", we ran the LDA again only on the news articles that belong to this topic to find the subtopics, totally 10 subtopics were identified. Out of these, we  selected "Corona Virus" as the topic for this study as it is the latest news articles. 

\section{Data Exploration}

The next step was to find the fake news articles, for this we injected a few synthetic fake news articles related to Corona virus based on the real news articles to the fake news dataset and then the  similarity was computed on the entire fake news dataset against the real news dataset to identify the relevant fake news articles. The traditional word frequency approaches does not perform well in capturing the semantic relationships between words. Hence prediction or probability-based approaches were developed to solve this problem. Mikolov et al. [10] proposed Continuous Bag Of Words (CBOW) a modified version of Bag Of Words (BOW) model, they found CBOW performed well in determining the semantic relationship between words in a given context when compared to the traditional word frequency approaches. In addition to CBOW, Mikolov et al., also proposed Continuous Skip-gram model, both of them fall under word2Vec (Word to Vector) architecture mainly used to measure the semantic relationship between words. All these new inventions in word embeddings gave birth to many new models which helped in encoding sentence and paragraphs, one such model is Universal sentence encoder based on deep neural networks architecture.Therefore, the Universal Sentence encoder was used [12] to generate sentence embeddings of the real news articles. Similarly, embeddings were generated for all the fake news articles. Later, the inner product of real and fake news embeddings were computed to determine the similarity and select the fake news articles that are similar to "Corona Virus" real news articles. At the last, the top 10 fake news articles were selected. 
Lastly, the Spacy library was used to tag the Parts Of Speech (POS) of the given news articles. Besides, the same library was used for NamedEntityRecognition (NER) to extract Person, Location, and Organization. POS and NER were used to select the required tokens based on the tags and generate the graphs as per the given topology shown in Figure 2. One sample instance of graph is shown in Table 1.

\section{Experiments}
Using the graph topology shown in Figure 2 as input to GBAD, we were able to discover several fake news.\\
Hardware specifications for all our experiments are as follows:
\begin{itemize}
\item Processor Intel(R) Core(TM) i3-5005U CPU @2.00GHz 2.00 GHz, 2 Core(s), 4 Logical Processor(s).
\item RAM 4.00GB.
\item Operating system: xubuntu 16.04.
\end{itemize}

\begin{figure}[tb]
 \centering 
  \framebox{\includegraphics[width=8.0cm,height=5.0cm]{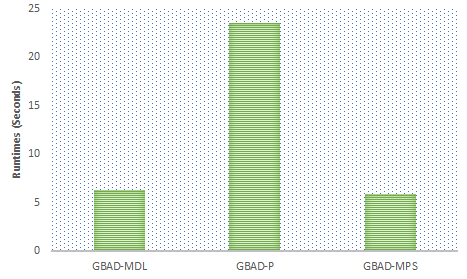}}
 \caption{Running times needed to discover patterns using GBAD}
 \label{fig:runtime}
\end{figure}

Running times needed to discover patterns using GBAD to process 3602 news instances can be seen in Figure \ref{fig:runtime}.

Run-time needed to discover patterns using GBAD-P is comparatively more than GBAD-MDL and GBAD-MPS.

\section{Results and Discussion}
\begin{figure}[!tb]
 \centering 
  \framebox{\includegraphics[width=8.0cm,height=5.0cm]{ 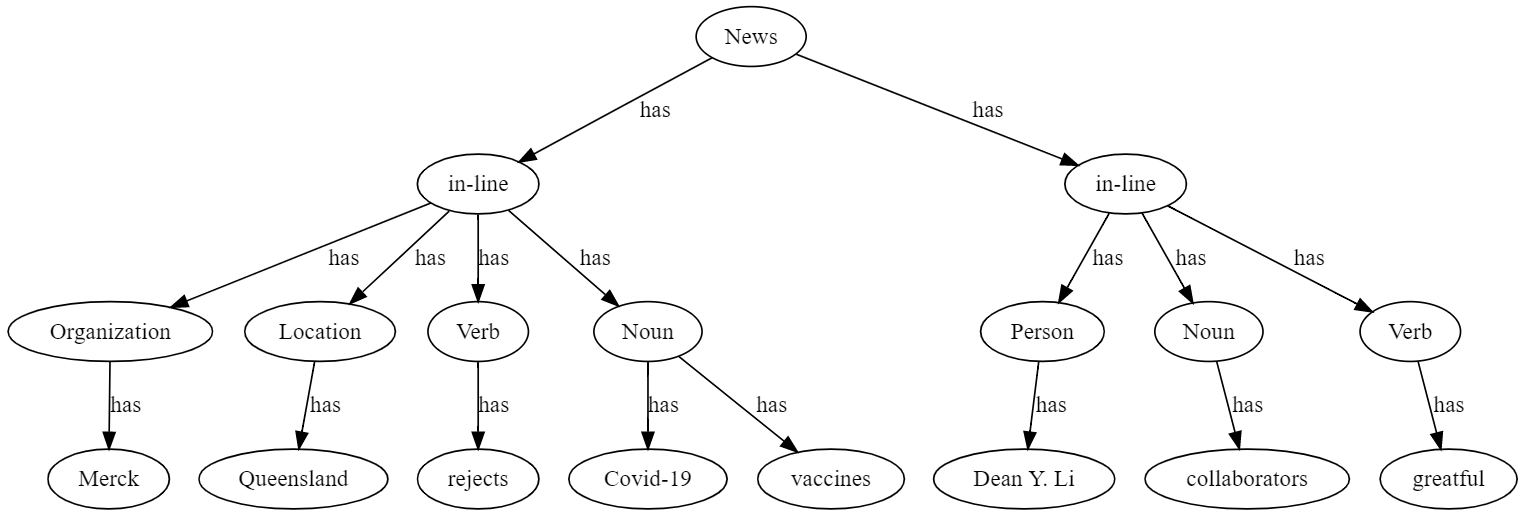}}
 \caption{Normative Pattern Found by GBAD-MDL}
 \label{fig:norm_mdl}
\end{figure}

\begin{figure}[!tb]
 \centering 
  \framebox{\includegraphics[width=8.0cm,height=5.0cm]{ 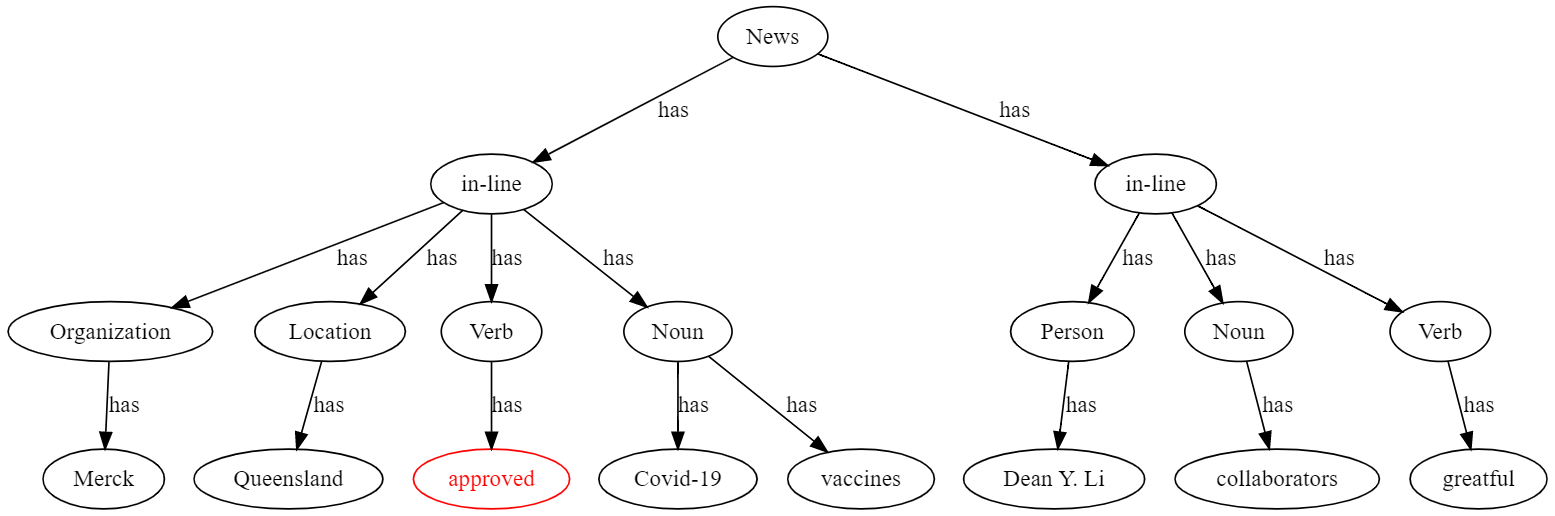}}
 \caption{Anomalous Pattern Found by GBAD-MDL}
 \label{fig:anom_mdl}
\end{figure}

\begin{figure}[!tb]
 \centering 
  \framebox{\includegraphics[width=8.0cm,height=5.0cm]{ 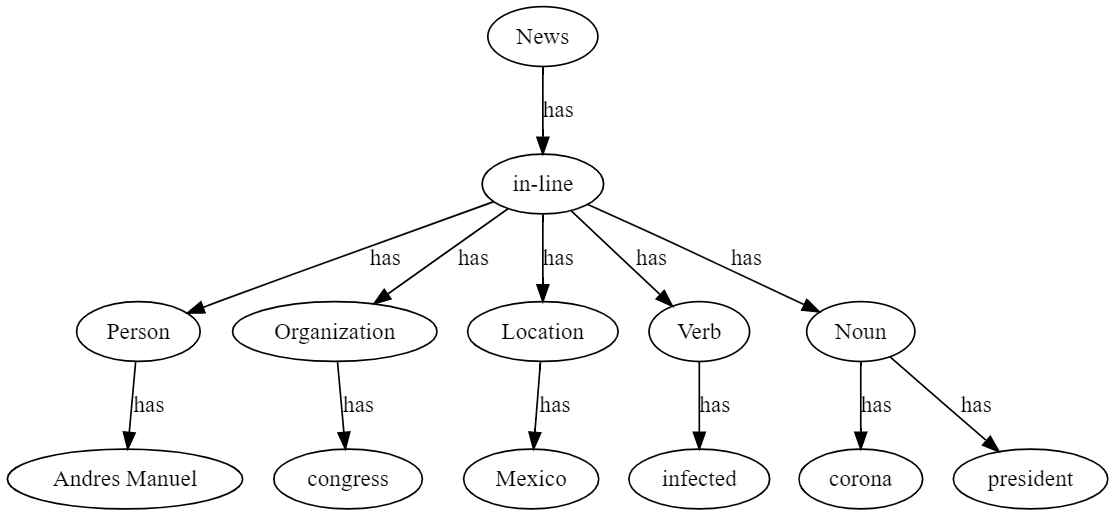}}
 \caption{Normative Pattern Found by GBAD-P}
 \label{fig:norm_gbadp}
\end{figure}

\begin{figure}[!tb]
 \centering 
  \framebox{\includegraphics[width=8.0cm,height=5.0cm]{ 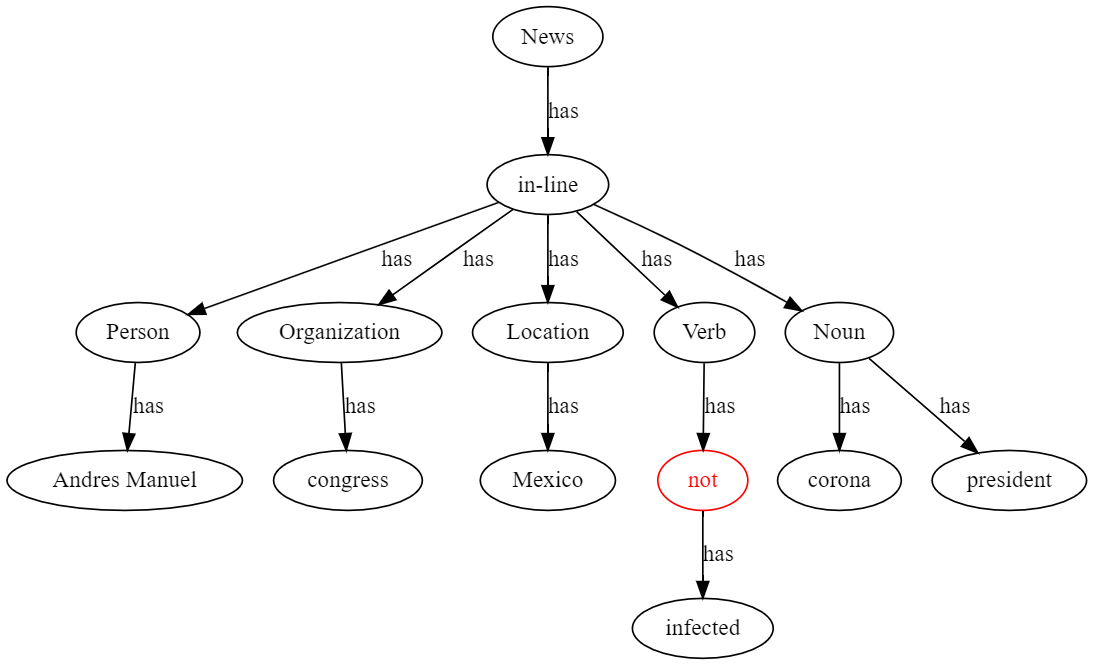}}
 \caption{Anomalous Pattern Found by GBAD-P}
 \label{fig:anom_gbadp}
\end{figure}

\begin{figure}[!tb]
 \centering 
  \framebox{\includegraphics[width=8.0cm,height=5.0cm]{ 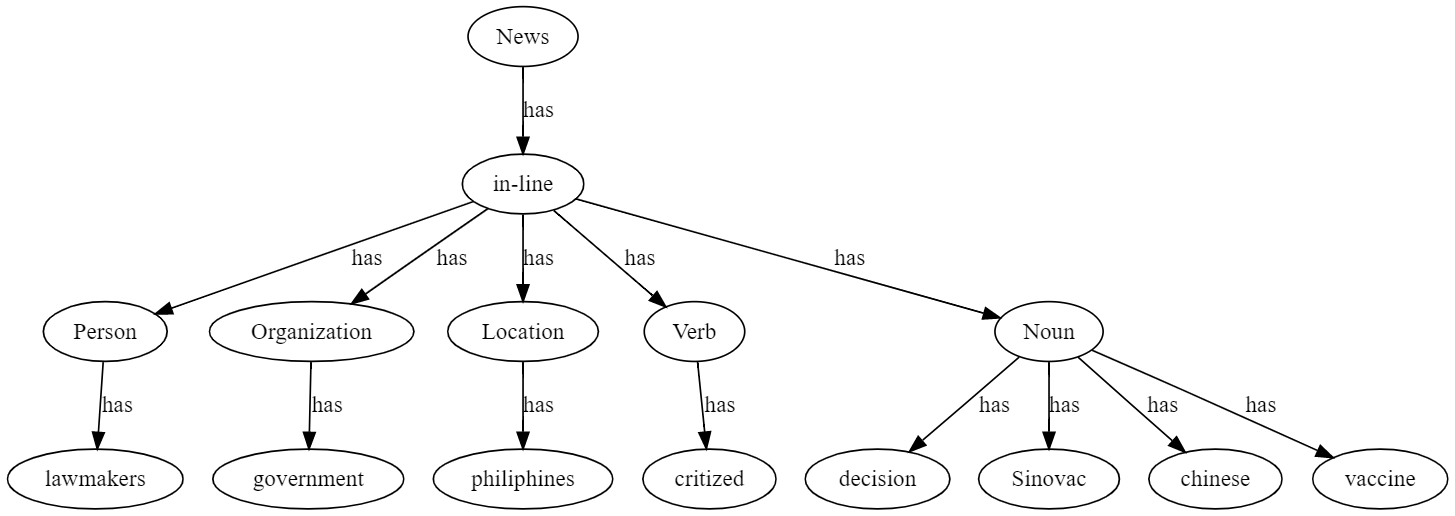}}
 \caption{Normative Pattern Found by GBAD-MPS}
 \label{fig:norm_mps}
\end{figure}

\begin{figure}[!tb]
 \centering 
  \framebox{\includegraphics[width=8.0cm,height=5.0cm]{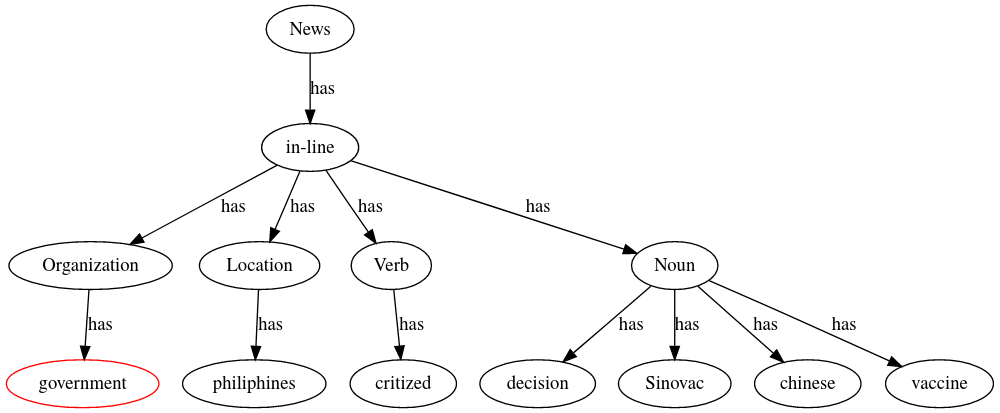}}
 \caption{Anomalous Pattern Found by GBAD-MPS}
 \label{fig:anom_mps}
\end{figure}

The GBAD algorithms discover anomalous substructures in the graph-representation of the provided data. 
\begin{itemize}
    \item Normative pattern and anomalous pattern found by GBAD-MDL algorithm can be seen in Figures \ref{fig:norm_mdl}, \ref{fig:anom_mdl}.From the normative pattern it is inferred that Covid vaccine is rejected. But the fake news found by GBAD-MDL is that the vaccine got approved.
    \item  Normative pattern and anomalous pattern found by GBAD-P algorithm can be seen in Figures \ref{fig:norm_gbadp}, \ref{fig:anom_gbadp}. From the normative pattern it is inferred that president is infected by corona. But the fake news found by GBAD-P is that the president is not infected by corona.
    \item  Normative pattern and anomalous pattern found by GBAD-MPS algorithm can be seen in Figures \ref{fig:norm_mps}, \ref{fig:anom_mps}.
    From the normative pattern it is inferred that lawmakers criticized Sinovac vaccine. But the fake news found by GBAD-MPS is that government criticized Sinovac vaccine.
\end{itemize}

\section{Conclusion and Future Work}
In this work, we used graph-based approach to discover contextual fake news articles. Using \textit{NLP} techniques, we are able to convert news articles to conceptual graphs. In this work, we analyzed only articles related to Corona, but in future we want to analyze varied domains and mine interesting patterns from heterogeneous data sets. In this work, we analyzed only English newspapers, whereas analyzing news articles of various language scan be a good future study. We also want to add some interactive visualization techniques that can provide quick insights and analytic.

\bibliographystyle{ACM-Reference-Format}

\end{document}